# Multilingual Topic Models for Unaligned Text

**Jordan Boyd-Graber**
35 Olden Street
Computer Science Dept.
Princeton University
Princeton, NJ 08540

**David M. Blei**
35 Olden Street
Computer Science Dept.
Princeton University
Princeton, NJ 08540

## Abstract

We develop the multilingual topic model for unaligned text (MuTo), a probabilistic model of text that is designed to analyze corpora composed of documents in two languages. From these documents, MuTo uses stochastic EM to simultaneously discover both a matching between the languages and multilingual latent topics. We demonstrate that MuTo is able to find shared topics on real-world multilingual corpora, successfully pairing related documents across languages. MuTo provides a new framework for creating multilingual topic models without needing carefully curated parallel corpora and allows applications built using the topic model formalism to be applied to a much wider class of corpora.

Topic models are a powerful formalism for unsupervised analysis of corpora [1, 8]. They are an important tool in information retrieval [27], sentiment analysis [25], and collaborative filtering [18]. When interpreted as a mixed membership model, similar assumptions have been successfully applied to vision [6], population survey analysis [4], and genetics [5].

In this work, we build on latent Dirichlet allocation (LDA) [2], a generative, probabilistic topic model of text. LDA assumes that documents have a distribution over topics and that these topics are distributions over the vocabulary. Posterior inference discovers the topics that best explain a corpus; the uncovered topics tend to reflect thematically consistent patterns of words [8]. The goal of this paper is to find topics that express thematic coherence across multiple languages.

LDA can capture coherence in a single language because semantically similar words tend to be used in similar contexts. This is not the case in multilingual corpora. For example, even though "Hund" and "hound" are orthographically similar and have nearly identical meanings in German and English (i.e., "dog"), they will likely not appear in similar contexts because almost all documents are written in a single language. Consequently, a topic model fit on a bilingual corpus reveals coherent topics but bifurcates the topic space between the two languages (Table 1). In order to build coherent topics across languages, there must be some connection to tie the languages together.

Previous multilingual topic models connect the languages by assuming parallelism at either the sentence level [28] or document level [13, 23, 19]. Many parallel corpora are available, but they represent a small fraction of corpora. They also tend to be relatively well annotated and understood, making them less suited for unsupervised methods like LDA. A topic model on unaligned text in multiple languages would allow the exciting applications developed for monolingual topics models to be applied to a broader class of corpora and would help monolingual users to explore and understand multilingual corpora.

We propose the MUltilingual TOpic model for unaligned text (MUTO). MUTO does not assume that it is given any explicit parallelism but instead discovers a parallelism at the vocabulary level. To find this parallelism, the model assumes that similar themes and ideas appear in both languages. For example, if the word "Hund" appears in the German side of the corpus, "hound" or "dog" should appear somewhere on the English side.

The assumption that similar terms will appear in similar contexts has also been used to build lexicons from non-parallel but comparable corpora. What makes contexts similar can be evaluated through such measures as cooccurrence [20, 24] or tf-idf [7]. Although the emphasis of our work is on building consistent topic spaces and not the task of building dictionaries *per se*, good translations are required to find consistent topics. However, we can build on successful techniques at building lexicons across languages.

This paper is organized as follows. We detail the model and its assumptions in Section 1, develop a stochastic expectation maximization (EM) inference procedure in Section 2, discuss the corpora and other linguistic resources necessary to evaluate the model in Section 3, and evaluate the performance of the model in Section 4.

## 1 Model

We assume that, given a bilingual corpus, similar themes will be expressed in both languages. If "dog," "bark," "hound," and "leash" are associated with a pet-related topic in English, we can find a set of pet-related words in German without having translated all the terms. If we can guess or we are told that "Hund" corresponds to one of these words, we can discover that words like "Leinen," "Halsband," and "Bellen" ("leash," "collar," and "bark," respectively) also appear with "Hund" in German, making it reasonable to guess that these words are part of the pet topic as expressed in German.

These steps—learning which words comprise topics within a language and learning word translations across languages—are both part of our model. In this section, we describe MUTO's generative model, first describing how a matching connects vocabulary terms across languages and then describing the process for using those matchings to create a multilingual topic model.

### 1.1 Matching across Vocabularies

We posit the following generative process to produce a bilingual corpus in a source language $S$ and a target language $T$. First, we select a matching $m$ over terms in both languages. The matching consists of pairs $(v_i, v_j)$ linking a term $v_i$ in the vocabulary of the first language $V_S$ to a term $v_j$ in the vocabulary of the second language $V_T$. A matching can be viewed as a bipartite graph with the words in one language $V_S$ on one side and $V_T$ on the other. A word is either unpaired or linked to a single node in the opposite language.

The use of a matching as a latent parameter is inspired by the matching canonical correlation analysis (MCCA) model [12], another method that induces a dictionary from arbitrary text. MCCA uses a matching to tie together words with similar meanings (where similarity is based on feature vectors representing context and morphology). We have a slightly looser assumption; we only require words with similar document level contexts to be matched. Another distinction is that instead of assuming a uniform prior over matchings, as in MCCA, we consider the matching to have a regularization term $\pi_{i,j}$ for each edge. We prefer larger values of $\pi_{i,j}$ in the matching.

This parameterization allows us to incorporate prior knowledge derived from morphological features, existing dictionaries, or dictionaries induced from non-parallel text. We can also use the knowledge gleaned from parallel corpora to understand the non-parallel corpus of interest. Sources for the matching prior $\pi$ are discussed in Section 3.

### 1.2 From Matchings to Topics

In MUTO, documents are generated conditioned on the matching. As in LDA, documents are endowed with a distribution over topics. Instead of being distributions over terms, topics in MUTO are distributions over pairs in $m$. Going back to our intuition, one such pair might be ("hund", "hound"), and it might have high probability in a pet-related topic. Another difference from LDA is that unmatched terms don't come from a topic but instead come from a unigram distribution specific to each language. The full generative process of the matching and both corpora follows:

1. Choose a matching $m$ where the probability of an edge $m_{i,j}$ being included is proportional to $\pi_{i,j}$
2. Choose multinomial term distributions:
   (a) For languages $L \in \{S, T\}$, choose background distributions $\rho_L \sim \text{Dir}(\gamma)$ over the words not in $m$.
   (b) For topic index $i = \{1, \ldots, K\}$, choose topic $\beta_i \sim \text{Dir}(\lambda)$ over the pairs $(v_S, v_T)$ in $m$.
3. For each document $d = \{1, \ldots D\}$ with language $l_d$:
   (a) Choose topic weights $\theta_d \sim \text{Dir}(\alpha)$.
   (b) For each $n = \{1, \ldots, M_d\}$ :
      i. Choose topic assignment $z_n \sim \text{Mult}(1, \theta_d)$.
      ii. Choose $c_n$ from {matched, unmatched} uniformly at random.
      iii. If $c_n$ matched, choose a pair $\sim \text{Mult}(1, \beta_{z_n}(m))$ and select the member of the pair consistent with $l_d$, the language of the document, for $w_n$.
      iv. If $c_n$ is unmatched, choose $w_n \sim \text{Mult}(1, \rho_{l_d})$.

Both $\rho$ and $\beta$ are distributions over words. The background distribution $\rho_S$ is a distribution over the $(|V_S| - |m|)$ words not in $m$, $\rho_T$ similarly for the other language, and $\beta$ is a distribution over the word pairs in $m$. Because a term is either part of a matching or not, these distributions partition the vocabulary.

| Topic 0 | Topic 1 | Topic 2 | Topic 3 |
|---|---|---|---|
| market | group | bericht | praesident |
| policy | vote | fraktion | menschenrecht |
| service | member | abstimmung | jahr |
| sector | committee | kollege | regierung |
| competition | report | ausschuss | parlament |
| system | matter | frage | mensch |
| employment | debate | antrag | hilfe |
| company | time | punkt | volk |
| union | resolution | abgeordnete | region |

Table 1: Four topics from a ten topic LDA model run on the German and English sections of Europarl. Without any connection between the two languages, the topics learned are language-specific.

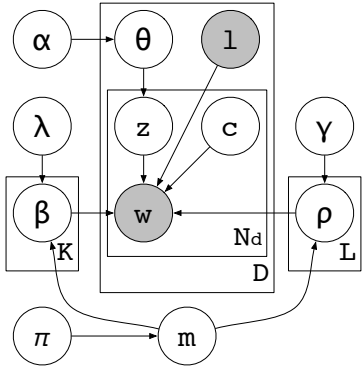

Figure 1: Graphical model for MUTO. The matching over vocabulary terms $m$ determines whether an observed word $w_n$ is drawn from a topic-specific distribution $\beta$ over matched pairs or from a language-specific background distribution $\rho$ over terms in a language.

The background distribution is the same for all documents. We choose not to have topic-specific distributions over unmatched words for two reasons. The first reason is to prevent topics from having divergent themes in different languages. For example, even if a topic had the matched pair ("Turkei", "Turkey"), distinct language topic multinomials over words could have "Istanbul," "Atatürk," and "NATO" in German but "stuffing," "gravy," and "cranberry" in English. The second reason is to encourage very frequent nouns that can be well explained by a language-specific distribution (and thus likely not to be topical) to remain unmatched.

## 2 Inference

Given two corpora, our goal is to infer the matching $m$, topics $\beta$, per-document topic distributions $\theta$, and topic assignments $z$. We solve this posterior inference problem with a stochastic EM algorithm [3]. There are two components of our inference procedure: finding the maximum a posteriori matching and sampling topic assignments given the matching.

We first discuss estimating the latent topic space given the matching. We use a collapsed Gibbs sampler [9] to sample the topic assignment of the $n^{th}$ word of the $d^{th}$ document conditioned on all other topic assignments and the matching, integrating over topic distributions $\beta$ and the document topic distribution $\theta$. $D_{d,i}$ is the number of words assigned to topic $i$ in document $d$ and $C_{i,t}$ is the number of times either of the terms in pair $t$ has been assigned topic $i$. For example, if $t = (\text{hund}, \text{hound})$, "hund" has been assigned topic three five times, and "hound" has been assigned topic three twice, then $C_{3,t} = 7$.

The conditional distribution for the topic assignment of matched words is

$$p(z_{d,n} = i | \boldsymbol{z}_{-i}, \boldsymbol{m}) \propto$$
$$\left(\frac{D_{d,i} + \frac{\alpha}{K}}{D_{d,\cdot} + \alpha}\right)\left(\frac{C_{i,m(w_n)} + \frac{\lambda}{|\boldsymbol{m}|}}{C_{i,\cdot} + \lambda}\right),$$

and unmatched words are assigned a topic based on the document topic assignments alone.

Now, we choose the maximum a posteriori matching given the topic assignments using the Hungarian algorithm [17]. We first consider how adding a single edge impacts the likelihood. Adding an edge $(i, j)$ means that the the occurrences of term $i$ in language $S$ and term $j$ in language $T$ come from the topic distributions instead of two different background distributions. So we must add the likelihood contribution of these new topic-specific occurrences to the likelihood and subtract the global language-multinomial contributions from the likelihood.

Using our posterior posterior estimates of topics $\beta$ and $\rho$ from the Markov chain, the number of times word $i$ appears in language $l$, $N_{l,i}$, and the combined topic count for the putative pair $C_{k,(i,j)}$, the resulting weight between term $i$ and term $j$ is

$$\mu_{i,j} = \sum_k C_{k,(i,j)} \log \beta_{k,(i,j)} \qquad (1)$$
$$- N_{S,i} \log \rho_{S,i} - N_{T,j} \log \rho_{T,j} + \log \pi_{i,j}.$$

Maximizing the sum of the weights included in our matching also maximizes the posterior probability of the matching.[1]

Intuitively, the matching encourages words to be paired together if they appear in similar topics, are not explained by the background language model, and are compatible with the preferences expressed by the matching prior $\pi_{i,j}$. The words that appear only in specialized contexts will be better modeled by topics rather than the background distribution.

MUTO requires an initial matching which can subsequently be improved. In all our experiments, the initial matching contained all words of length greater than five characters that appear in both languages. For languages that share similar orthography, this produces a high precision initial matching [16].

This model suffers from overfitting; running stochastic EM to convergence results in matchings between words that are unrelated. We correct for overfitting by stopping inference after three M steps (each stochastic E step used 250 Gibbs sampling iterations) and gradually increasing the size of the allowed matching after each iteration, as in [12]. Correcting for overfitting in a more principled way, such as by explicitly controlling the number of matchings or employing a more expressive prior over the matchings, is left for future work.

---
[1] Note that adding a term to the matching also potentially changes the support of $\beta$ and $\rho$. Thus, the counts associated with terms $i$ and $j$ appear in the estimate for both $\beta$ (corresponding to the log likelihood contribution if the match is included) and $\rho$ (corresponding to the log likelihood if the match is not added); this is handled by the Gibbs sampler across M-step updates because the topic assignments alone represent the state.

## 3 Data

We studied MUTO on two corpora with four sources for the matching prior. We use a matching prior term $\pi$ in order to incorporate prior information about which matches the model should prefer. Which source is used depends on how much information is available for the language pair of interest.

**Pointwise Mutual Information from Parallel Text** Even if our dataset of interest is not parallel, we can exploit information from available parallel corpora in order to formulate $\pi$. For one construction of $\pi$, we computed the pointwise mutual information (PMI) for terms appearing in the translation of aligned sentences in a small German-English news corpus [14].

**Dictionary** If a machine readable dictionary is available, we can use the existence of a link in the dictionary as our matching prior. We used the Ding dictionary [21]; terms with $N$ translations were given weight $\frac{1}{N}$ with all of the possible translations given in the dictionary (connections which the dictionary did not admit were effectively disallowed). This gives extra weight to unambiguous translations.

**Edit Distance** If there are no reliable resources for our language pair but we assume there is significant borrowing or morphological similarity between the languages, we can use string similarity to formulate $\pi$. We used

$$\pi_{i,j} = \frac{1}{0.1 + \text{ED}(v_i, v_j)}.$$

Although deeper morphological knowledge could be encoded using a specially derived substitution penalty, all substitutions and deletions were penalized equally in our experiments.

**MCCA** For a bilingual corpus, matching canonical correlation analysis model finds a mapping from latent points $z_i, z_j \in \mathbb{R}^n$ to the observed feature vector $f(v_i)$ for a term $v_i$ in one language and $f(v_j)$ for a term $v_j$ in the second language. We run the MCCA algorithm on our bilingual corpus to learn this mapping and use

$$\log \pi_{i,j} \approx -||z_i - z_j||.$$

This distance between preimages of feature vectors in the latent space is proportional to the weight used in MCCA algorithm to construct matchings. We used the same method for selecting an initial matching for MCCA as for MUTO. Thus, identical pairs were used as the initial seed matching rather than randomly selected pairs from a dictionary. When we used MCCA as a prior, we ran MCCA on the same dataset as a first step to compute the prior weights.

### 3.1 Corpora

Although MUTO is designed with non-parallel corpora in mind, we use parallel corpora in our experiments for the purposes of evaluation. We emphasize that the model does not use the parallel structure of the corpus. Using parallel corpora also guarantees that similar themes will be discussed, one of our key assumptions.

First, we analyzed the German and English proceedings of the European Parliament [15], where each chapter is considered to be a distinct document. Each document on the English side of the corpus has a direct translation on the German side; we used a sample of 2796 documents.

Another corpus with more variation between languages is Wikipedia. A bilingual corpus with explicit mappings between documents can be assembled by taking Wikipedia articles that have cross-language links between the German and English versions. The documents in this corpus have similar themes but can vary considerably. Documents often address different aspects of the same topic (e.g. the English article will usually have more content relevant to British or American readers) and thus are not generally direct translations as in the case of the Europarl corpus. We used a sample of 2038 titles marked as German-English equivalents by Wikipedia metadata.

We used a a part of speech tagger [22] to remove all non-noun words. Because nouns are more likely to be constituents of topics [10] than other parts of speech, this ensures that terms relevant to our topics will still be included. It also prevents uninformative but frequent terms, such as highly inflected verbs, from being included in the matching.[2] The 2500 most frequent terms were used as our vocabulary. Larger vocabulary sizes make computing the matching more difficult as the full weight matrix scales as $V^2$, although this could be addressed by filtering unlikely weights.

## 4 Experiments

We examine the performance of MUTO on three criteria. First, we examine the qualitative coherence of learned top-ics, which provides intuition about the workings of the model. Second, we assess the accuracy of the learned matchings, which ensures that the topics that we discover are not built on unreasonable linguistic assumptions. Last, we investigate the extent to which MUTO can recover the parallel structure of the corpus, which emulates a document retrieval task: given a query document in the source language, how well can MUTO find the corresponding document in the target language?

---

[2] Although we used a part of speech tagger for filtering, a stop word filter would yield a similar result if a tagger or part of speech dictionary were unavailable.

In order to distinguish the effect of the learned matching from the information already available through the matching prior $\pi$, for each model we also considered a "prior only" version where the matching weights are held fixed and the matching uses only the prior weights (i.e., only $\pi_{i,j}$ is used in Equation 2).

### 4.1 Learned Topics

To better illustrate the latent structure used by MUTO and build insight into the workings of the model, Table 2 shows topics learned from German and English articles in Wikipedia. Each topic is a distribution over pairs of terms from both languages, and the topics seem to demonstrate a thematic coherence. For example, Topic 0 is about computers, Topic 2 concerns science, etc.

Using edit distance as a matching prior allowed us to find identical terms that have similar topic profiles in both languages such as "computer," "lovelace," and "software." It also has allowed us to find terms like "objekt," "astronom," "programm," and "werk" that are similar both in terms of orthography and topic usage.

Mistakes in the matching can have different consequences. For instance, "earth" is matched with "stickstoff" (nitrogen) in Topic 2. Although the meanings of the words are different, they appear in sufficiently similar science-oriented contexts that it doesn't harm the coherence of the topic.

In contrast, poor matches can dilute topics. For example, Topic 4 in Table 2 seems to be split between both math and Roman history. This encourages matches between terms like "rome" in English and "römer" in German. While "römer" can refer to inhabitants of Rome, it can also refer to the historically important Danish mathematician and astronomer of the same name. This combination of different topics is further reinforced in subsequent iterations with more Roman / mathematical pairings.

Spurious matches accumulate over time, especially in the version of MUTO with no prior. Table 3 shows how poor matches lead to a lack of correspondence between topics across languages. Instead of developing independent, internally coherent topics in both languages (as was observed in the naïve LDA model in Table 1), the arbitrary matches pull the topics in many directions, creating incoherent top-ics and incorrect matches.

| Topic 0 | Topic 1 |
|---|---|
| wikipedia:agatha | alexander:temperatur |
| degree:christie | country:organisation |
| month:miss | city:leistung |
| director:hercule | province:mcewan |
| alphabet:poirot | empire:auftreten |
| issue:marple | asia:factory |
| ocean:modern | afghanistan:status |
| atlantic:allgemein | roman:auseinandersetzung |
| murder:harz | government:verband |
| military:murder | century:fremde |

Table 3: Two topics from a twenty topic MUTO model trained on Wikipedia with no prior on the matching. Each topic is a distribution over pairs; the top pairs from each topic are shown. Without appropriate guidance from the matching prior, poor translations accumulate and topics show no thematic coherence.

### 4.2 Matching Translation Accuracy

Given a learned matching, we can ask what percentage of the pairs are consistent with a dictionary [21]. This gives an idea of the consistency of topics at the vocabulary level.

These results further demonstrate the need to influence the choice of matching pairs. Figure 2 shows the accuracy of multiple choices for computing the matching prior. If no matching prior is used, essentially no correct matches are chosen.

Models trained on Wikipedia have lower vocabulary accuracies than models trained on Europarl. This reflects a broader vocabulary, a less parallel structure, and the limited coverage of the dictionary. For both corpora, and for all prior weights, the accuracy of the matchings found by MUTO is nearly indistinguishable from matchings induced by using the prior weights alone. Adding the topic structure neither hurts nor helps the translation accuracy.

### 4.3 Matching Documents

While translation accuracy measures the quality of the matching learned by the algorithm, how well we recover the parallel document structure of the corpora measures the quality of the latent topic space MUTO uncovers. Both of our corpora have explicit matches between documents across languages, so an effective multilingual topic model should associate the same topics with each document pair regardless of the language.

We compare MUTO against models on bilingual corpora that do not have a matching across languages: LDA applied to a multilingual corpus using a *union* and *intersection* vocabulary. For the *union* vocabulary, all words from both languages are retained and the language of documents is ignored. Posterior inference in this setup effectively parti-tions the topics into topics for each language, as in Table 1. For the *intersection* vocabulary, the language of the document is ignored, but all terms in one language which don't have an identical counterpart in the other are removed.

| Topic 0 | Topic 1 | Topic 2 | Topic 3 | Topic 4 |
|---|---|---|---|---|
| **apple:apple** | **nbsp:nbsp** | **bell:bell** | **lincoln:lincoln** | **quot:quot** |
| **code:code** | pair:jahr | **nobel:nobel** | **abraham:abraham** | time:schatten |
| **anime:anime** | exposure:kategorie | **alfred:alfred** | **union:union** | world:kontakt |
| **computer:computer** | space:sprache | claim:ampere | united:nationale | history:roemisch |
| **style:style** | bind:bild | **alexander:alexander** | **president:praesident** | **number:nummer** |
| **character:charakter** | price:thumb | proton:graham | **party:partei** | math:with |
| **ascii:ascii** | belt:zeit | telephone:behandlung | states:status | term:zero |
| **line:linie** | decade:bernstein | **experiment:experiment** | state:statue | **axiom:axiom** |
| **program:programm** | deal:teil | invention:groesse | republican:mondlandung | **system:system** |
| **software:software** | **name:name** | acoustics:strom | **illinois:illinois** | **theory:theorie** |

Table 2: Five topics from a twenty topic MUTO model trained on Wikipedia using edit distance as the matching prior $\pi$. Each topic is a distribution over pairs; the top pairs from each topic are shown. Topics display a semantic coherence consistent with both languages. Correctly matched word pairs are in bold.

If MUTO finds a consistent latent topic space, then the distribution of topics $\theta$ for matched document pairs should be similar. For each document $d$, we computed the the Hellinger distance between its $\theta$ and all other documents' $\theta$ and ranked them. The proportion of documents less similar to $d$ than its designated match measures how consistent our topics are across languages. These results are presented in Figure 3.

For a truly parallel corpus like Europarl, the baseline of using the intersection vocabulary did very well (because it essentially matched infrequent nouns). On the less parallel Wikipedia corpus, the intersection baseline did worse than all of the MUTO methods. On both corpora, the union baseline did little better than random guessing.

Although morphological cues were effective for finding high-accuracy matchings, this information doesn't necessarily match documents well. The edit weight prior on Wikipedia worked well because the vocabulary of pages varies substantially depending on the subject, but methods that use morphological features (edit distance and MCCA) were not effective on the more homogenous Europarl corpus, performing little better than chance.

Even by themselves, our matching priors do a good job of connecting words across the languages' vocabularies. On the Wikipedia corpus, all did better than the LDA baselines and MUTO without a prior. This suggests that an end-user interested in obtaining a multilingual topic model could obtain acceptable results by simply constructing a matching using one of the schemes outlined in Section 3 and running MUTO using this static matching.

However, MUTO can perform better if the matchings are allowed to adjust to reflect the data. For many conditions, MUTO with the matchings updated using the weights in Equation 2 performs better on the document matching task than using the matching prior alone.

## 5 Discussion

In this work, we presented MUTO, a model that simultaneously finds topic spaces and matchings in multiple languages. In evaluations on real-world data, MUTO recovers matched documents better than the prior alone. This suggests that MUTO can be used as a foundation for multilingual applications using the topic modeling formalism and as an aid in corpus exploration.

Corpus exploration is especially important for multilingual corpora, as users are often more comfortable with one language in a corpus than the other. Using a more widely used language such as English or French to provide readable signposts, multilingual topic models could help uncertain readers find relevant documents in the language of interest.

MUTO makes no linguistic assumptions about the input data that precludes finding relationships and semantic equivalences on symbols from other discrete vocabularies. Data are often presented in multiple forms; models that can explicitly learn the relationships between different modalities could help better explain and annotate pairings of words and images, words and sound, genes in different organisms, or metadata and text.

Conversely, adding more linguistic assumptions such as incorporating local syntax in the form of feature vectors is an effective way to find translations without using parallel corpora. Using such local information within MUTO, rather than just as a prior over the matching, would allow the quality of translations to improve and would be another alternative to the techniques that attempt to combine local context with topic models [26, 11].

With models like MUTO, we can remove the assumption of monolingual corpora from topic models. Exploring this new latent topic space also offers new opportunities for researchers interested in multilingual corpora for machine translation, linguistic phylogeny, and semantics.

## 6 Acknowledgements

The authors would like to thanks Aria Haghighi and Percy Liang for providing code and advice. Conversations with Richard Socher and Christiane Fellbaum were invaluable in developing this model. David M. Blei is supported by ONR 175-6343, NSF CAREER 0745520, and grants from Google and Microsoft.

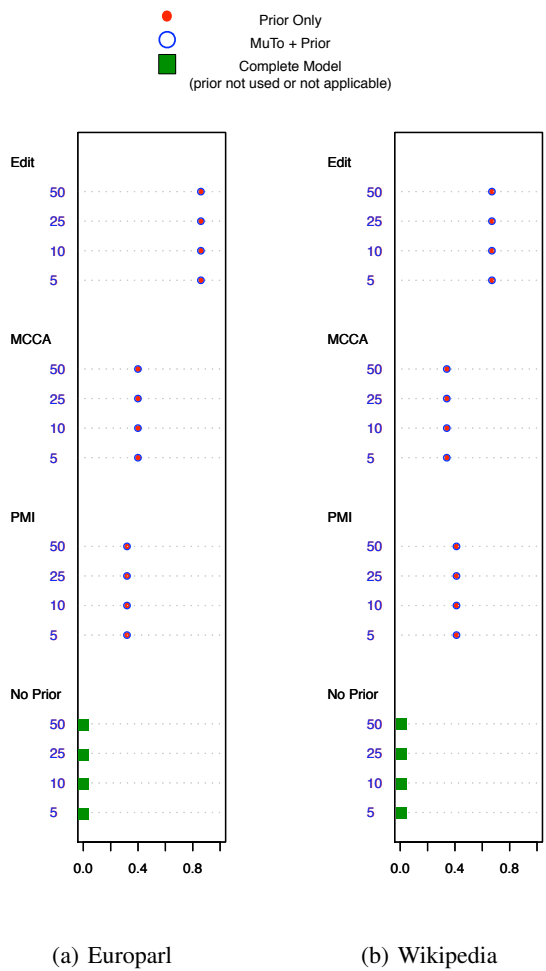
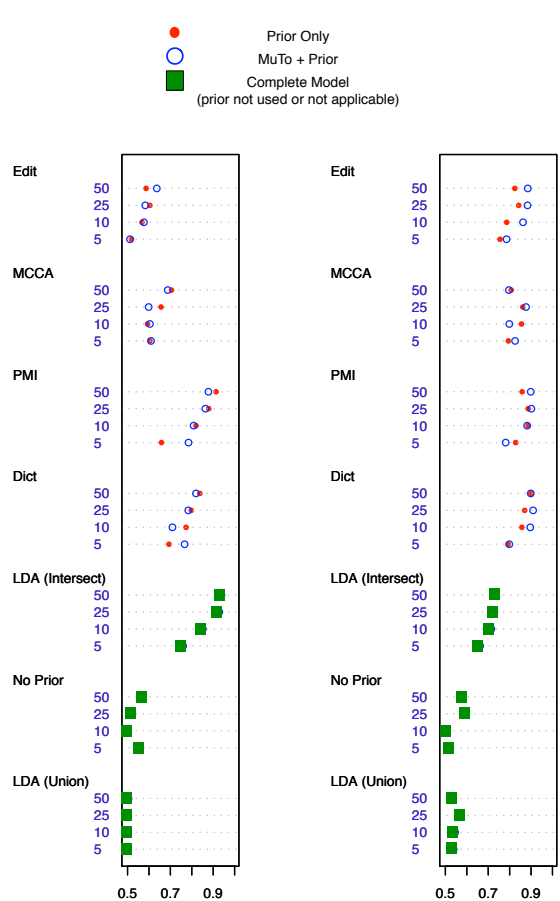

Figure 2: Each group corresponds to a method for computing the weights used to select a matching; each group has values for 5, 10, 25, and 50 topics. The x-axis is the percentage of terms where a translation was found in a dictionary. Where applicable, for each matching prior source, we compare the matching found using MUTO with a matching found using only the prior. Because this evaluation used the Ding dictionary [21], the matching prior derived from the dictionary is not shown.

Figure 3: Each group corresponds to a method for creating a matching prior $\pi$; each group has values for 5, 10, 25, and 50 topics. The full MUTO model is also compared to the model that uses the matching prior alone to select the matching. The x-axis is the proportion of documents whose topics were less similar than the correct match across languages (higher values, denoting fewer misranked documents, are better).